\documentclass[conference]{IEEEtran}
\IEEEoverridecommandlockouts
\usepackage{cite}
\usepackage{amsmath,amssymb,amsfonts}
\usepackage{graphicx}
\usepackage{textcomp}
\usepackage{tabularx}
\usepackage{threeparttable}
\usepackage[colorlinks=true, allcolors=black]{hyperref}

\begin{document}

\title{FASA: Feedback-Aware Sampling Adaptation for Efficient Diffusion-Based VLA Models
\thanks{This work was supported by Shenzhen Key Industry Research and Development Program (No. ZDCY20250901100302003), Shenzhen-Hong Kong Joint Funding Project (Category A) (No. SGDX20240115103359001) and the General Program of Guangdong Basic and Applied Basic Research Foundation (No. 2026A1515010273).}
}

\author{\IEEEauthorblockN{1\textsuperscript{st} Yuchen Han}
\IEEEauthorblockA{\textit{South China University of Technology}\\
Guangzhou, China \\
Ping An Technology (Shenzhen) Co., Ltd.\\
Shenzhen, China \\
csychan@mail.scut.edu.cn}
\and
\IEEEauthorblockN{2\textsuperscript{nd} Jianhan Wu\textsuperscript{\textdagger}}
\IEEEauthorblockA{\textit{Ping An Technology (Shenzhen) Co., Ltd.}\\
Shenzhen, China \\
jerric\_wu@163.com\\
\textsuperscript{\textdagger}Corresponding Author}
\and
\IEEEauthorblockN{3\textsuperscript{rd} Xiaoyang Qu}
\IEEEauthorblockA{\textit{Ping An Technology (Shenzhen) Co., Ltd.}\\
Shenzhen, China \\
quxiaoy@gmail.com}
\and
\IEEEauthorblockN{4\textsuperscript{th} Lingwei Kong}
\IEEEauthorblockA{\textit{Ping An Technology (Shenzhen) Co., Ltd.}\\
Shenzhen, China \\
jsjyklw@outlook.com}
\and
\IEEEauthorblockN{5\textsuperscript{th} Shiyi Li}
\IEEEauthorblockA{\textit{Harbin Institute of Technology}\\
Shenzhen, China \\
lishiyi@hit.edu.cn}
\and
\IEEEauthorblockN{6\textsuperscript{th} Jianzong Wang}
\IEEEauthorblockA{\textit{Ping An Technology (Shenzhen) Co., Ltd.}\\
Shenzhen, China \\
jzwang@188.com}
}

\IEEEpubid{979-8-3195-9980-3/26/\$31.00~\copyright~2026 IEEE}

\maketitle

\begin{abstract}
Diffusion-based Vision-Language-Action (VLA) models achieve strong performance in embodied tasks, but their iterative sampling imposes heavy computational and memory-access cost, blocking real-time deployment on edge platforms. Existing acceleration methods either require expensive training (e.g., distillation, flow matching) or degrade perception via statically scheduled pruning and caching, ignoring the dynamic workload variance of robotic interactions. This paper presents FASA (\underline{F}eedback-\underline{A}ware \underline{S}ampling \underline{A}daptation), a training-free runtime framework that treats real-time multimodal feedback as a control signal for the denoising pipeline: an interaction-driven range adaptor modulates the global sampling-step budget based on visual and gripper-force feedback, and a proprioception-aware step adaptor pinpoints the optimized step within the adapted range. This co-designed framework allows the underlying hardware architecture to adaptively match the workload demands of different execution phases. Comparative evaluations across several benchmarks show that the inference speed can be increased by up to 1.45$\times$ while maintaining competitive success rates, providing a novel dynamic runtime architecture paradigm for deploying heavy generative embodied AI workloads onto resource-constrained computing platforms.
\end{abstract}


\section{Introduction}
\label{sec:intro}

Vision-Language-Action (VLA) models integrate visual perception, language understanding, and action generation, enabling robots to interpret multimodal instructions and execute physical tasks \cite{liuRDT1BDiffusionFoundation2024,linEvo1LightweightVisionLanguageAction2026}. Among policy architectures, diffusion-based action heads excel at modeling multi-modal action distributions through iterative denoising \cite{chiDiffusionPolicyVisuomotor2024}. However, executing these workloads requires hundreds of denoising iterations per control step, incurring latency and energy costs that hinder real-time edge deployment.

Existing acceleration efforts fall into two categories. \textit{Training-based} methods, such as consistency distillation \cite{prasadConsistencyPolicyAccelerated2024}, reduce sampling steps but require costly retraining and transfer poorly across policies. \textit{Training-free} methods prune visual tokens, cache perception features \cite{yangEfficientVLATrainingFreeAcceleration2025,xuVLACacheEfficientVisionLanguageAction2025}, or exit language-model layers early \cite{yueDeeRVLADynamicInference2024}. However, their reductions follow fixed schedules or input-level statistics rather than the precision actually demanded by the ongoing interaction, weakening perception in precision-critical phases while still overspending compute on trivial motions. Moreover, none of them exploits the real-time feedback stream that embodied platforms uniquely provide to control the depth of the diffusion loop.

This feedback stream is a control resource unique to embodied computing: robotic tasks alternate between precision-critical phases (e.g., contact-rich manipulation) and compute-tolerant phases (e.g., coarse reaching), and no static sampling budget can match. We therefore propose \textbf{FASA} (\textbf{F}eedback-\textbf{A}ware \textbf{S}ampling \textbf{A}daptation), a training-free, plug-and-play runtime framework that treats environmental and proprioceptive feedback as control signals for the denoising pipeline; since the number of denoising steps directly determines action precision \cite{chiDiffusionPolicyVisuomotor2024}, FASA dynamically allocates this budget at runtime. An \textit{Interaction-Driven Sampling Range Adaptation} module fuses visual-feature variation and gripper-force feedback to scale the sampling-step boundaries, and a \textit{Proprioception-Aware Sampling Step Adaptation} module computes a kinematic activity metric from joint states to pinpoint an optimized step within that range. We implement FASA on 3D Diffuser Actor, Diffusion Policy, and RDT, and deploy a force-free variant on Evo-1, yielding \textbf{1.40$\times$} on RLBench, up to \textbf{1.45$\times$} on ManiSkill3, and \textbf{1.23$\times$} on LIBERO with competitive success rates.

\IEEEpubidadjcol

\begin{figure*}[tbp]
  \centering
  \includegraphics[width=0.85\linewidth]{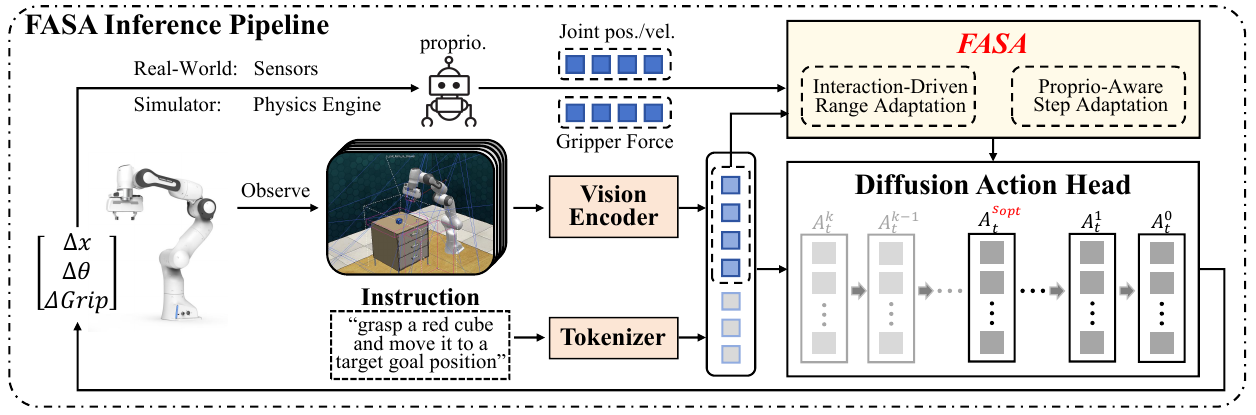}
 \caption{\textbf{FASA Inference Pipeline}. FASA integrates proprioceptive and visual feedback to determine an optimized sampling step $s_{\text{opt}}$. \textit{Proprio-Aware} = Proprioception-Aware; \textit{proprio.} = proprioception; \textit{pos./vel.} = joint positions/velocities.}
  \label{fig:overview}
\end{figure*}

\begin{figure*}[tbp]
  \centering
  \includegraphics[width=0.85\linewidth]{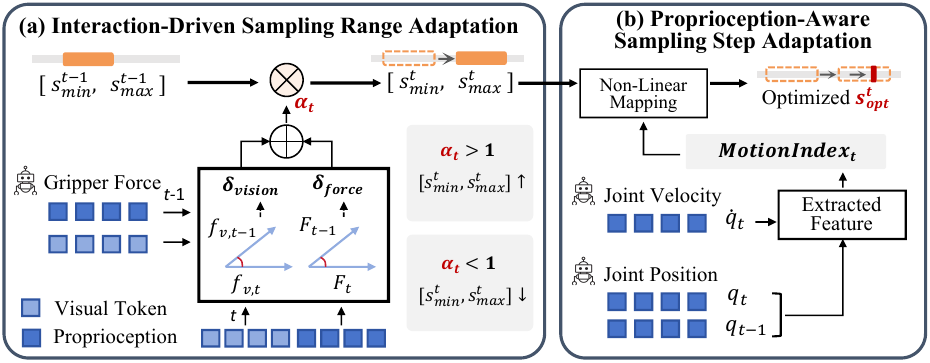}
  \caption{\textbf{FASA Framework}. \textbf{(a)} Interaction-Driven Sampling Range Adaptation takes visual feature $f_{v,t}$ and gripper force $F_t$ to compute a modulation factor $\alpha_t$ that adjusts the range $[s_{\min}^t, s_{\max}^t]$. \textbf{(b)} Proprioception-Aware Sampling Step Adaptation extracts joint position $q_t$ and velocity $\dot{q}_t$ to form $\text{MotionIndex}_t$; both determine the optimized sampling step $s_{\text{opt}}^t$.}
  \label{fig:details}
\end{figure*}

\section{Method}



A standard diffusion-based policy denoises actions with a fixed $K$-step loop at every execution step. FASA replaces this fixed budget with feedback-conditioned scheduling, shown in Fig.~\ref{fig:overview}.

\begin{table*}[htb]
    \centering
    \caption{\label{tab:rlbench}Evaluation of FASA on RLBench (18 tasks)}
    \begin{tabularx}{\textwidth}{l|*{10}{>{\centering\arraybackslash}X}}
        \hline
        \textbf{Method} & \textbf{Avg. (\%)$\uparrow$}                           & \textbf{Speedup$\uparrow$}                              & \begin{tabular}[c]{@{}c@{}}close\\ jar\end{tabular}     & \begin{tabular}[c]{@{}c@{}}insert\\ peg\end{tabular}  & \begin{tabular}[c]{@{}c@{}}light\\ bulb\end{tabular} & \begin{tabular}[c]{@{}c@{}}meat\\ grill\end{tabular}  & \begin{tabular}[c]{@{}c@{}}open\\ drawer\end{tabular}  & \begin{tabular}[c]{@{}c@{}}sort\\ shape\end{tabular} & \begin{tabular}[c]{@{}c@{}}place\\ cup\end{tabular}  & \begin{tabular}[c]{@{}c@{}}place\\ wine\end{tabular} \\ \hline
        3D-DA           & 75.96                                                  & 1.00$\times$                                            & 30.67                                                   & 62.67                                                 & \textbf{73.33}                                       & 96.00                                                 & \textbf{92.00}                                         & 45.33                                                & 14.67                                                & \textbf{93.33}                                       \\
        FASA            & \textbf{77.48}                                         & \textbf{1.40$\times$}                                   & \textbf{42.67}                                          & \textbf{70.67}                                        & 69.33                                                & \textbf{96.00}                                        & 89.33                                                  & \textbf{49.33}                                       & \textbf{21.33}                                       & 90.67                                                \\ \hline
        \textbf{Method} & \begin{tabular}[c]{@{}c@{}}push\\ buttons\end{tabular} & \begin{tabular}[c]{@{}c@{}}put in\\ cupbd.\end{tabular} & \begin{tabular}[c]{@{}c@{}}put in\\ drawer\end{tabular} & \begin{tabular}[c]{@{}c@{}}put in\\ safe\end{tabular} & \begin{tabular}[c]{@{}c@{}}drag\\ stick\end{tabular} & \begin{tabular}[c]{@{}c@{}}slide\\ block\end{tabular} & \begin{tabular}[c]{@{}c@{}}stack\\ blocks\end{tabular} & \begin{tabular}[c]{@{}c@{}}stack\\ cups\end{tabular} & \begin{tabular}[c]{@{}c@{}}sweep\\ dust\end{tabular} & \begin{tabular}[c]{@{}c@{}}turn\\ tap\end{tabular}   \\ \hline
        3D-DA           & 97.33                                                  & 81.33                                                   & \textbf{94.67}                                          & \textbf{100.0}                                        & 100.0                                                & \textbf{98.67}                                        & 54.00                                                  & \textbf{53.33}                                       & 81.33                                                & \textbf{98.67}                                       \\
        FASA            & \textbf{97.33}                                         & \textbf{82.67}                                          & 93.33                                                   & 96.00                                                 & \textbf{100.0}                                       & 97.33                                                 & \textbf{66.67}                                         & 49.33                                                & \textbf{85.33}                                       & 97.33                                                \\ \hline 
    \end{tabularx}

    \begin{tablenotes}
        \footnotesize
        \item Task success rates and average action generation speedups of FASA (with \textbf{3D} \textbf{D}iffuser \textbf{A}ctor as baseline) are reported. \textit{Avg.} denotes the average task success rate across all 18 tasks. \textit{Speedup$\uparrow$} denotes the ratio of baseline time over FASA time.
    \end{tablenotes}
\end{table*}

\begin{table}[tbp]
    \centering
    \caption{\label{tab:maniskill3}Evaluation of FASA on ManiSkill3 (3 tasks)}
    \newcolumntype{Y}{>{\centering\arraybackslash}X}
    \begin{tabularx}{\columnwidth}{l|YYY|YY}
        \hline
        \textbf{Method} & \begin{tabular}[c]{@{}c@{}}pick\\ cube\end{tabular} & \begin{tabular}[c]{@{}c@{}}push\\ cube\end{tabular} & \begin{tabular}[c]{@{}c@{}}stack\\ cube\end{tabular} & \textbf{Avg.(\%)$\uparrow$} & \textbf{Speedup$\uparrow$} \\ \hline
        DP              & 52.50                                               & 69.17                                               & 89.17                                                & 70.28                                                                  & 1.00$\times$               \\
        FASA            & \textbf{64.17}                                      & \textbf{69.17}                                      & \textbf{95.83}                                       & \textbf{76.39}                                                         & \textbf{1.45$\times$}      \\ \hline
        RDT             & 79.17                                               & 100.0                                               & 79.17                                                & 86.11                                                                  & 1.00$\times$               \\
        FASA            & \textbf{80.00}                                      & \textbf{100.0}                                      & \textbf{82.50}                                       & \textbf{87.50}                                                         & \textbf{1.23$\times$}      \\ \hline
    \end{tabularx}

    \begin{tablenotes}
        \footnotesize
        \item Success rates and inference speedups of FASA (with \textbf{D}iffusion \textbf{P}olicy and \textbf{RDT} as baselines) are shown for each task, with Mean Accuracy and average Speedup across all tasks. This table shares metrics with Table~\ref{tab:rlbench}.
    \end{tablenotes}
\end{table}

\subsection{Interaction-Driven Sampling Range Adaptation}
\label{sec:interaction-driven}

FASA first utilizes interaction information to dynamically adjust the range of sampling steps $[s_{\min}, s_{\max}]$. We quantify the temporal variation of the interaction state as:
\begin{equation}
    \label{eq:delta}
    \begin{array}{c}
    \delta_{\text{force}}^t= -\cos (F_{t}, F_{t-1}),  \\
    \delta_{\text{vision}}^t = - \cos(f_{v,t}, f_{v, t-1} ),
    \end{array}
\end{equation}
where $F_{t}$ and $f_{v,t}$ denote the gripper force and visual feature at execution step $t$, respectively. $F_{t}$ is provided by the robot's sensors or the physics engine in simulation, while $f_{v,t}$ is extracted from the current observation using the VLA model's original vision encoder, aligning FASA's visual feedback with the policy's representation. The two variation scores are fused into a modulation factor:
\begin{equation}
    \label{eq:alpha}
    \alpha_t = \frac{1}{2} \cdot \delta_{\text{vision}}^t + \frac{1}{2} \cdot \delta_{\text{force}}^t + \beta.
\end{equation}
Both $\delta$ terms are bounded cosine-distance scores in $[-1,1]$, so equal weighting keeps the fusion scale-consistent without modality-specific tuning. The bias $\beta=1.5$ acts as a linear rectifier: since the variation term approaches $-1$ during smooth, quasi-static motion, $\beta$ guarantees a positive minimum multiplier ($\alpha_t \approx 0.5$), preventing generation collapse. Hence $\alpha_t < 1$ compresses the execution path during compute-tolerant stable phases, while $\alpha_t > 1$ allocates a larger arithmetic budget for dynamic interactions. The sampling range is updated at each execution step as:
\begin{equation}
    \label{eq:sampling_range}
    \begin{array}{c}
    s_{\min}^{t} = s_{\min}^{t-1} \cdot \alpha_t,\quad
    s_{\max}^{t} = s_{\max}^{t-1} \cdot \alpha_t,
    \end{array}
\end{equation}
where the initialization satisfies $s_{\max}^0 = 2 s_{\min}^0$. In implementation, the updated bounds are smoothed by an exponential moving average and clipped to a valid range to avoid abrupt oscillation.

\subsection{Proprioception-Aware Sampling Step Adaptation}
\label{sec:proprio-aware}

While the range adaptor determines the global sampling budget, the concrete sampling step must align with the robot's real-time physical dynamics. We define a kinematic activity metric:
\begin{equation}
    \label{eq:motion_index}
    \text{MotionIndex}_{t} =
    \lVert \dot{q}_{t} \rVert  +
    \lVert q_{t} - q_{t-1} \rVert,
\end{equation}
where $\dot{q}_{t}$ is the instantaneous joint velocity and $\lVert q_{t} - q_{t-1} \rVert$ is the realized positional displacement. Both terms are computed from the normalized proprioceptive state already consumed by the policy backbone (joint positions and velocities standardized per joint), so $\text{MotionIndex}_{t}$ is dimensionless and the two terms are directly commensurable. A higher $\text{MotionIndex}_{t}$ signifies a highly dynamic phase that necessitates finer temporal resolution. The metric is mapped to the optimized sampling step via a non-linear gating mechanism:
\begin{equation}
    \label{eq:optimized_step}
    s_{\text{opt}}^{t} = s_{\min}^{t} + (s_{\max}^{t} - s_{\min}^{t}) \cdot
    \sigma(\text{MotionIndex}_{t}),
\end{equation}
where $\sigma(\cdot)$ denotes the sigmoid function. FASA thus realizes feedback-driven control of diffusion inference without training.

\section{Experiments}

\subsection{Experimental Setup}
\label{sec:experimental_setup}

\textbf{Policies and benchmarks.}
We evaluate 3D Diffuser Actor (3D-DA) \cite{ke3DDiffuserActor2025} on RLBench \cite{jamesRLBenchRobotLearning2020}, for which we use 18 language-conditioned tabletop tasks with multi-view vision and proprioception; Diffusion Policy (DP)\footnote{DP is a visuomotor policy rather than a VLA model, but it is the foundational diffusion action architecture widely adopted by VLA models.} \cite{chiDiffusionPolicyVisuomotor2024} and RDT \cite{liuRDT1BDiffusionFoundation2024} on ManiSkill3 \cite{taoDemonstratingGPUParallelized2025}, which is a GPU-parallelized simulator for contact-rich manipulation; and Evo-1 \cite{linEvo1LightweightVisionLanguageAction2026} on LIBERO \cite{liuLIBEROBenchmarkingKnowledge2023a}, for which we use 40 language-conditioned tasks from four suites. All comparisons follow the experimental settings of their corresponding baseline policies. For ManiSkill3, we select the three best-performing RDT baseline tasks---\textit{PickCube}, \textit{PushCube}, and \textit{StackCube}---from the five released tasks. A force-free FASA variant is used on LIBERO because the environment exposes no gripper-force readings. We report task success rate and action-generation speedup, defined as the ratio of baseline to FASA action-generation time.

\textbf{Platforms.}
We evaluate 3D-DA and Evo-1 on an NVIDIA RTX 4070 Laptop GPU with 8\,GB VRAM, and DP and RDT on an NVIDIA RTX 3090 GPU with 24\,GB VRAM.

\textbf{Implementation.}
FASA is a lightweight, hardware-agnostic runtime module that intercepts the policy inference loop. Its per-step overhead comprises an additional invocation of the policy's vision encoder, two cosine similarities, and the MotionIndex norms, as described in Sec.~\ref{sec:interaction-driven}. The required proprioceptive signals are universally accessible through standard physics-engine APIs or robot control interfaces, so FASA requires no modification to the robot's control stack.

\begin{table}[tbp]
    \centering
    \caption{\label{tab:libero}Evaluation of force-free FASA on LIBERO (40 tasks)}
    \small
    \newcolumntype{Y}{>{\centering\arraybackslash}X}
    \begin{tabularx}{\columnwidth}{l|YYYYY}
        \hline
        \textbf{Method} & \textbf{Avg.(\%)} & Spatial & Object & Goal & Long \\ \hline
        Evo-1 $\uparrow$ & 93.0 & \textbf{91.0} & 98.0 & 92.0 & 91.0 \\
        FASA $\uparrow$ & \textbf{93.5} & 87.0 & 98.0 & \textbf{94.0} & \textbf{95.0} \\ \hline
        Evo-1 Lat. $\downarrow$ & 233.71 & 234.70 & 233.74 & 233.54 & 232.88 \\
        FASA Lat. $\downarrow$ & \textbf{190.61} & \textbf{195.01} & \textbf{196.59} & \textbf{198.05} & \textbf{172.78} \\ \hline
        Speedup $\uparrow$ & \textbf{1.23$\times$} & \textbf{1.20$\times$} & \textbf{1.19$\times$} & \textbf{1.18$\times$} & \textbf{1.35$\times$} \\ \hline
    \end{tabularx}
    
    \begin{tablenotes}
        \footnotesize
        \item Success rate and action-generation latency are reported for Evo-1 and FASA across 40 LIBERO tasks. \textit{Lat.} is the average action-generation time per request in milliseconds. Speedup is calculated as the ratio of Evo-1 latency to FASA latency.
    \end{tablenotes}
\end{table}



\subsection{Main Results}

\textbf{RLBench.}
Table~\ref{tab:rlbench} shows that FASA raises the average success rate from 75.96\% to 77.48\% while achieving a 1.40$\times$ action-generation speedup. Large gains on \textit{close jar}, 30.67\%$\rightarrow$42.67\%, and \textit{stack blocks}, 54.00\%$\rightarrow$66.67\%, demonstrate that adaptive sampling improves efficiency without sacrificing task execution quality.

\textbf{ManiSkill3.}
FASA improves DP from 70.28\% to 76.39\% with a 1.45$\times$ speedup and RDT from 86.11\% to 87.50\% with a 1.23$\times$ speedup. These gains across distinct policies demonstrate FASA's transferability, while the smaller speedup on 10-step RDT shows that acceleration headroom scales with the baseline sampling budget.

\textbf{LIBERO.}
Force-free FASA raises the overall success rate from 93.0\% to 93.5\% while achieving a 1.23$\times$ speedup as the average denoising budget falls from 32 to 19.79 steps. \textit{LIBERO-Long} reaches 1.35$\times$ speedup with a four-point success-rate gain, whereas the four-point drop on \textit{LIBERO-Spatial} indicates that fine-grained spatial control requires a more conservative schedule. These results establish that feedback-adaptive acceleration remains effective without force sensing.

\begin{figure}[tbp]
  \centering
  \includegraphics[width=0.95\linewidth]{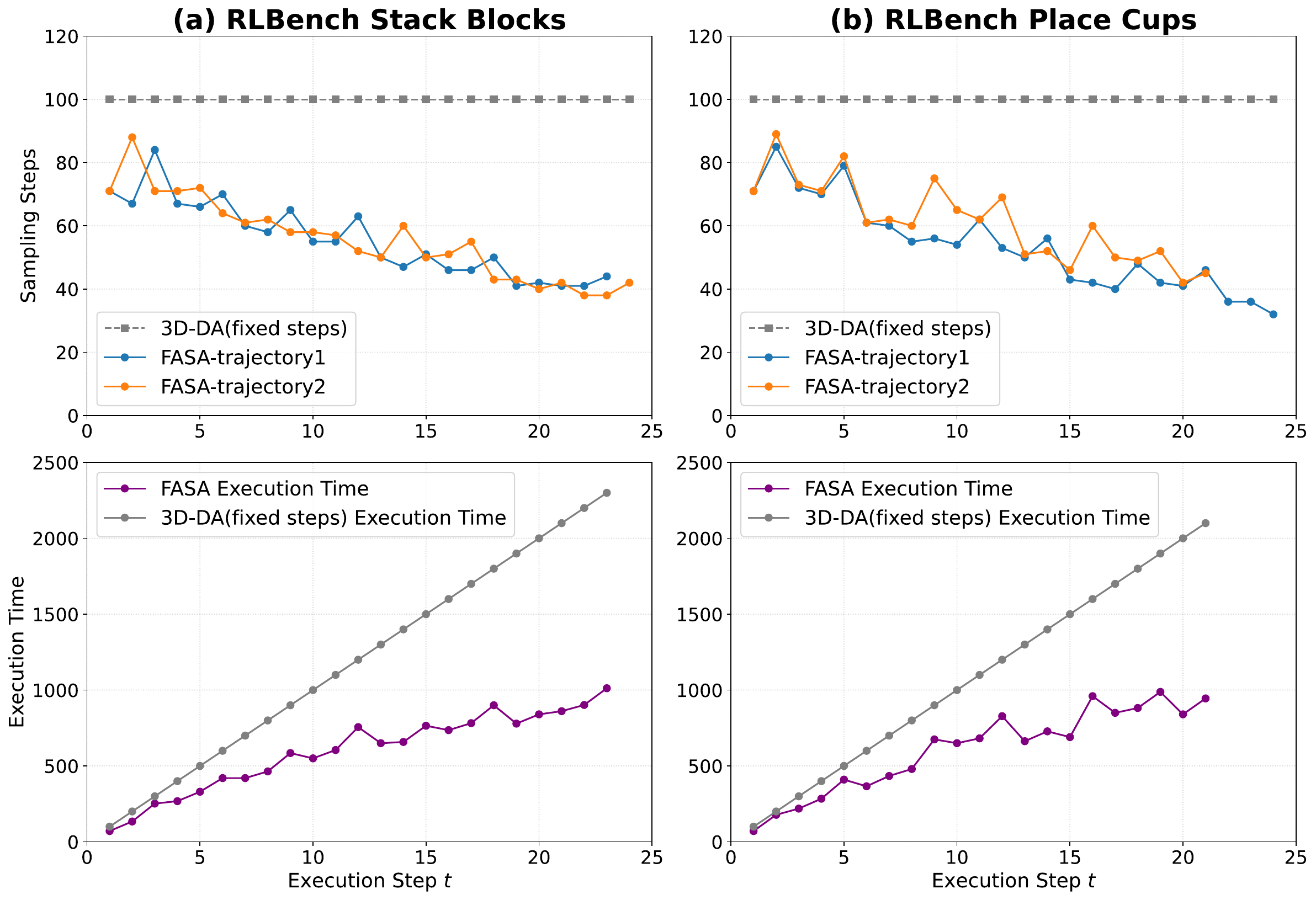}
  \caption{\textbf{Visualization of FASA's Adaptive Sampling}. Two successful RLBench cases (\textit{Stack Blocks} and \textit{Place Cups}), showing sampling steps and execution time.}
  \label{fig:visualization}
\end{figure}

Fig.~\ref{fig:visualization} visualizes the sampling steps of two successful RLBench cases. The sampling steps exhibit a downward trend while the context and proprioceptive state remain stable, and escalate when feedback signals significant state transitions, confirming closed-loop dynamic regulation.

\begin{table}[tbp]
    \centering
    \caption{\label{tab:ablation}Ablation Study on FASA}
    \small
    \newcolumntype{Y}{>{\centering\arraybackslash}X}
    \begin{tabularx}{\columnwidth}{l|YYY}
    \hline
    Method            & Acc.(\%)$\uparrow$ & Time(s)$\downarrow$ & Speedup$\uparrow$     \\ \hline
    3D-DA             & 77.22            & 33.59               & 1.00$\times$          \\
    w/ Range Adaptor  & 76.33            & \textbf{20.52}      & \textbf{1.64$\times$} \\
    w/ Step Adaptor & 77.06            & 28.46               & 1.18$\times$          \\
    FASA      & \textbf{77.50}   & 23.64               & 1.42$\times$          \\ \hline
    \end{tabularx}

    \begin{tablenotes}
        \footnotesize
        \item \textit{Range/Step Adaptor}: the two FASA modules in Secs.~\ref{sec:interaction-driven}/\ref{sec:proprio-aware}; 3D-DA: baseline without either.
    \end{tablenotes}
\end{table}

\subsection{Ablation Study}

Table~\ref{tab:ablation} analyzes the contributions of the two modules. In the Range-Adaptor-only variant, the Step Adaptor is disabled and the policy samples at the current upper bound $s_{\max}^{t}$ of the adapted range at every execution step. Since the range contracts globally under stable feedback, this variant attains the highest speedup (1.64$\times$) but slightly compromises accuracy. The Step Adaptor alone performs fine-grained adjustments that preserve precision, yet yields only 1.18$\times$. FASA couples global range modulation with local step refinement, retaining a 1.42$\times$ speedup with the highest success rate (77.50\%), confirming that both levels of adaptation are essential.

\section{Conclusion}

We presented FASA, a training-free runtime framework that adapts the number of denoising steps of diffusion-based VLA inference to real-time visual, force, and proprioceptive feedback. Without training or modifying model weights, FASA achieves up to 1.45$\times$ speedup on RLBench, ManiSkill3, and LIBERO while maintaining or improving success rates, establishing feedback-driven compute scheduling for generative embodied-AI workloads on resource-constrained platforms. Future work includes real-robot validation in force-free settings and combining FASA with token-level acceleration.


\bibliographystyle{IEEEbib}
\bibliography{references}

\end{document}